\documentclass[conference]{IEEEtran}

\makeatletter

\def\ps@IEEEtitlepagestyle{%
  \def\@oddfoot{\mycopyrightnotice}%
  \def\@evenfoot{}%
}
\def\mycopyrightnotice{%
  {\footnotesize \hfill}% <--- Change here
  \gdef\mycopyrightnotice{}
}

\usepackage{blindtext}
\usepackage{eso-pic}
\IEEEoverridecommandlockouts
\usepackage{amsmath,amssymb,amsfonts}
\usepackage{algorithmic}
\usepackage{graphicx}
\usepackage{caption}
\usepackage{textcomp}
\usepackage{xcolor}

\usepackage{placeins}
\usepackage{booktabs}
\usepackage{url}
\usepackage{balance}

\usepackage{hyperref}
\usepackage{academicons} % or \usepackage{orcidlink} if preferred
\usepackage{xcolor}      % needed for icon color
\usepackage{orcidlink}
\usepackage{cite}

\def\BibTeX{{\rm B\kern-.05em{\sc i\kern-.025em b}\kern-.08em
    T\kern-.1667em\lower.7ex\hbox{E}\kern-.125emX}}
    
\usepackage{eso-pic}
\newcommand\AtPageUpperMyright[1]{\AtPageUpperLeft{%
 \put(\LenToUnit{0.17\paperwidth},\LenToUnit{-2cm}){%
     \parbox{0.9\textwidth}{\raggedleft\fontsize{8}{11}\selectfont #1}}%
 }}%
\newcommand{\conf}[1]{%
\AddToShipoutPictureBG*{%
\AtPageUpperMyright{#1}
}
}

\begin{document}
\title{\vspace*{1cm} Optimizing FPGA and Wafer Test Coverage with Spatial Sampling and Machine Learning\\
%{\footnotesize \textsuperscript{*}Note: Sub-titles are not captured in Xplore and should not be used}
}

\author{\IEEEauthorblockN{1\textsuperscript{st} Wang WeiQuan}
\IEEEauthorblockA{\textit{Interdisciplinary Faculty of Science and Engineering} \\
\textit{Shimane University}\\
Matsue, Japan \\
n24m119@matsu.shimane-u.ac.jp}
\and
\IEEEauthorblockN{2\textsuperscript{nd} Riaz-ul-Haque Mian}
\IEEEauthorblockA{\textit{Interdisciplinary Faculty of Science and Engineering} \\
\textit{Shimane University}\\
Matsue, Japan \\
riaz@cis.shimane-u.ac.jp}
}

\maketitle
\conf{\textit{This paper has been accepted for oral presentation at the 5th International Conference on Electrical, Computer and Energy Technologies (ICECET 2025), 3–6 July 2025, Paris, France.}}

\begin{abstract}
%Testing remains a major cost driver in semiconductor manufacturing.  To cut test counts without sacrificing accuracy, three baseline sampling schemes—Random, Stratified and \emph{k}-means—were evaluated on industrial wafer and silicon-FPGA data.  Building on these, \textbf{Short Distance Elimination (SDE)} has been introduced, a spatial-diversity filter that discards samples located too close to those already chosen.  Embedding SDE into Stratified and \emph{k}-means yields two hybrids: S-SDE and K-SDE.  With Gaussian Process Regression as the predictor and an optimized distance threshold of \((\alpha,\beta)=(2,2)\), the hybrids cut root-mean-square deviation (RMSD) by up to 16\% on wafers and 13\% on FPGAs compared with their respective baselines.  These results show that enforcing spatial dispersion during sampling is a simple yet effective lever for reducing test effort while preserving predictive fidelity.

In semiconductor manufacturing, testing costs remain significantly high, especially during wafer and FPGA testing. To reduce the number of required tests while maintaining predictive accuracy, this study investigates three baseline sampling strategies: Random Sampling, Stratified Sampling, and k-means Clustering Sampling. To further enhance these methods, this study proposes a novel algorithm that improves the sampling quality of each approach. This research is conducted using real industrial production data from wafer-level tests and silicon measurements from various FPGAs. This study introduces two hybrid strategies: Stratified with Short Distance Elimination (S-SDE) and k-means with Short Distance Elimination (K-SDE). Their performance is evaluated within the framework of Gaussian Process Regression (GPR) for predicting wafer and FPGA test data. At the core of our proposed approach is the Short Distance Elimination (SDE) algorithm, which excludes spatially proximate candidate points during sampling, thereby ensuring a more uniform distribution of training data across the physical domain\cite{zhang21}. A parameter sweep was conducted over the $(\alpha, \beta)$ thresholds—where $\alpha, \beta \in \{0, 1, 2, 3, 4\}$ and not both zero—to identify the optimal combination that minimizes RMSD. Experimental results on a randomly selected wafer file reveal that $(\alpha,\beta) = (2,2)$ yields the lowest RMSD. Accordingly, all subsequent experiments adopt this parameter configuration. The experimental results demonstrate that the proposed SDE-based strategies enhance predictive accuracy: K-SDE improves upon k-means sampling by 16.26\% (wafer) and 13.07\% (FPGA), while S-SDE improves upon stratified sampling by 16.49\% (wafer) and 8.84\% (FPGA).

\end{abstract}

%\copyrightnotice{XXX-X-XXXX-XXXX-X/XX/\$XX.00 ©20XX IEEE}


\section{Introduction}
Over the past several decades, the semiconductor manufacturing industry has experienced rapid development, yet the costs associated with testing have risen concurrently \cite{rasmussen06}. In producing high-precision and high-complexity devices such as wafers and FPGAs, performing a complete physical test on every device or node is both time consuming and expensive---requiring advanced test equipment and significant human resources. In today’s highly competitive market, reducing testing expenses while maintaining high yield and product quality is a major challenge\cite{lee20,li18}.

To address this challenge, a new paradigm---"partial measurement + machine learning prediction"---has emerged\cite{nguyen19}. In this strategy, only a subset (e.g., 10\%) of the full dataset is physically tested, while the remaining 90\% is predicted using a trained machine learning model\cite{chen21}. Gaussian Process Regression (GPR) has gained widespread acceptance for wafer and FPGA test modeling due to its advantages in dealing with small sample sizes, noisy data, and uncertainty quantification \cite{riaz24,riaz01,bishop06}. Prior research has demonstrated that GPR effectively captures the spatial variation common in semiconductor processes, adapting well to defect patterns on wafers and the distribution of FPGA paths.

However, if the 10\% training data is not selected appropriately, model performance will deteriorate. Although traditional random sampling is simple, it is susceptible to imbalance\cite{garcia19}. Stratified sampling and k-means clustering sampling---which segment or cluster based on measurement values (e.g., dynamic current or oscillator frequency)---can improve representativeness but typically overlook the spatial distribution\cite{li18}. This may lead to the selection of nearby points, reducing overall spatial coverage.

To overcome these limitations, the Short Distance Elimination (SDE) sampling strategy is proposed. SDE excludes candidate points that are too close in the spatial domain, ensuring a more widely dispersed set of training points. Hybrid strategies are formed by combining this spatial strategy with traditional methods: S-SDE (Stratified + SDE) and K-SDE (k-means + SDE). A systematic experimental evaluation of Random, Stratified, k-means, S-SDE, and K-SDE sampling strategies is then performed.

The key contributions of this paper are summarized as follows:

\begin{enumerate}
    \item A systematic comparison of three baseline sampling strategies—Random Sampling, Stratified Sampling, and k-means Clustering Sampling—for wafer and FPGA test prediction using Gaussian Process Regression (GPR) is presented.
    
    \item A novel Short Distance Elimination (SDE) algorithm is proposed that enhances spatial diversity in the training data by eliminating closely located candidate points during sampling.
    
    \item Two hybrid strategies, Stratified + SDE (S-SDE) and k-means + SDE (K-SDE), are introduced to combine value-based partitioning with spatial dispersion, improving both representativeness and spatial coverage.
    
    \item Experimental results show that sampling methods based on measurement values (Stratified and k-means) outperform pure random sampling in predictive accuracy\cite{roy21}.
    
    \item Incorporating spatial dispersion through SDE further improves performance. The proposed K-SDE method achieves the best overall accuracy on both industry wafer data and silicon FPGA datasets.
    
    \item The proposed SDE-based strategies significantly improve prediction accuracy: K-SDE improves upon k-means by 16.26\% (wafer) and 13.07\% (FPGA), while S-SDE improves upon stratified sampling by 16.49\% (wafer) and 8.84\% (FPGA).
\end{enumerate}

The rest of the paper is organized as follows:

Section~\ref{Related_Work} reviews related literature, including Gaussian Process Regression (GPR) applications and limitations of existing sampling methods in semiconductor testing. Section~\ref{Methodology} outlines the overall methodology, including the mathematical formulation of GPR and the implementation details of Random, Stratified, and k-means clustering-based sampling strategies. Section~\ref{sec:SDE} introduces the proposed Short Distance Elimination (SDE) method, along with its hybrid variants S-SDE and K-SDE. This section also includes algorithm descriptions and sampling logic. Section~\ref{sec:result} presents experimental results on real-world wafer and FPGA datasets. Performance is evaluated using RMSD, and the effectiveness of each sampling method is compared. Visualizations such as spatial heatmaps and performance graphs are also included. Finally, Section~\ref{sec:conclusion} concludes the paper with a summary of findings and outlines future research directions, including the application of SDE to other machine learning models and real-time sampling environments.

\section{Related Work} \label{sec:related}
\label{Related_Work}

\subsection{Gaussian Process Regression in Semiconductor Testing}
Gaussian Process Regression (GPR) is a Bayesian nonparametric approach for function modeling that is particularly effective in small-sample and noisy environments \cite{rasmussen06}. In semiconductor testing, GPR employs kernel functions to model the relationship between spatial coordinates and measurement values~\cite{mian2025custom,zhao21}, capturing complex spatial and global trends. For example, Riaz-ul-Haque \textit{et al.} \cite{riaz24} applied GPR for multi-site RF integrated circuit testing, successfully capturing wafer-level spatial variations, while Bishop \cite{bishop06} demonstrated its robustness in noisy and nonlinear datasets. These studies justify the adoption of GPR as the primary predictive model for wafer and FPGA test data in the present study.

\subsection{Sampling Methods and Their Limitations}
\begin{enumerate}
    \item \textbf{Random Sampling:} Random sampling is simple but may inadvertently over-sample certain regions and neglect extreme or critical areas, leading to prediction errors\cite{garcia19}.
    \item \textbf{Stratified Sampling:} By dividing measurement values into different strata based on quantiles or specific thresholds, stratified sampling ensures balanced representation across data ranges \cite{chen18,li18}. However, it ignores the spatial coordinates of wafer\cite{park20} and FPGA data.
    \item \textbf{k-means Clustering Sampling:} Utilizing the k-means algorithm \cite{jain10} partitions the data into clusters representing different modes of the measurement values.\cite{kumar21,li21} While this method can better handle multimodal distributions, it similarly neglects spatial dispersion.
\end{enumerate}

\subsection{Incorporating Spatial Information via Short Distance Elimination (SDE)}
Various grid-based and distance-based screening strategies have been proposed to address the spatial clustering of sampling points \cite{riaz24,das20,singh21,xu21}. The SDE method introduced in this paper excludes candidate points that lie within a predetermined threshold distance from already selected samples, ensuring more uniform spatial coverage. This spatial sampling can be further integrated with stratification or k-means clustering to form the hybrid methods S-SDE and K-SDE, thereby simultaneously enhancing numerical representativeness and spatial dispersion\cite{zhang21}.

\section{Methodology} \label{sec:methodology}
\label{Methodology}
The objective of this study is to extract 10\% of the measurements from large wafer or FPGA datasets for training purposes and to use GPR to predict the remaining 90\% of the data. The mathematical foundation of GPR, the individual sampling strategies, and the formalization of the hybrid methods S-SDE and K-SDE are described below.

\subsection{Gaussian Process Regression (GPR)}

\subsubsection{Basic Formulation}
A Gaussian Process (GP) assumes that the function values \( f(\mathbf{x}_i) \) for any finite set of input points are jointly Gaussian distributed \cite{rasmussen06}:
\begin{equation}
f(\mathbf{x}) \sim \mathcal{GP}\Bigl(m(\mathbf{x}),\, k(\mathbf{x}, \mathbf{x}')\Bigr),
\end{equation}
where \( m(\mathbf{x}) \) is the mean function (often set to zero) and \( k(\mathbf{x}, \mathbf{x}') \) is the covariance kernel function.

\subsubsection{RBF Kernel}
The Radial Basis Function (RBF) kernel is adopted:
\begin{equation}
k\Bigl(\mathbf{x}_i, \mathbf{x}_j\Bigr) 
= \sigma_f^2 \exp\!\Biggl(- \frac{\|\mathbf{x}_i - \mathbf{x}_j\|^2}{2l^2}\Biggr),
\end{equation}
where \( \mathbf{x}_i = (x_i, y_i) \), \( l \) is the length-scale, and \( \sigma_f^2 \) is the signal variance.

\subsubsection{Training and Prediction}
Given the training set \( \mathcal{D}_{\text{train}} \) of size \( M \) with observed values \( \mathbf{v} \in \mathbb{R}^M \) and kernel matrix \( \mathbf{K} \in \mathbb{R}^{M\times M} \) (with noise variance \( \sigma_n^2 \)), the predictive mean \( \hat{f}_* \) for a test point \( \mathbf{x}_* \) is given by:
\begin{equation}
\hat{f}_* = \mathbf{k}_*^T \Bigl(\mathbf{K} + \sigma_n^2 \mathbf{I}\Bigr)^{-1} \mathbf{v},
\end{equation}
where \( \mathbf{k}_* \) is the vector of kernel similarities between \( \mathbf{x}_* \) and each training point.

\subsection{Sampling Strategies}
Let the dataset be represented as:
\begin{equation}
\mathcal{D}=\{(x_i, y_i, v_i)\}_{i=1}^N.
\end{equation}
Ten percent (denoted by \( p = 0.1 \)) of the data are selected for training the GPR model, while the remaining 90\% are used for testing. The following sampling methods are considered:

\subsubsection{Random Sampling}
The training set is obtained by randomly choosing \( pN \) data points from \( \mathcal{D} \):
\begin{equation}
\mathcal{D}_{\text{train}} = \mathrm{RandomChoose}\!\bigl(\mathcal{D},\, p\bigr).
\end{equation}

\subsubsection{Stratified Sampling}
\paragraph{Stratification Definition:}  
Measurement values \( v_i \) are divided into \( S \) strata based on quantiles or predetermined thresholds \( \{b_1, \dots, b_{S-1}\} \):
\begin{equation}
\text{Stratum}_s = \{(x_i, y_i, v_i) \mid b_{s-1} \le v_i < b_s\}.
\end{equation}
\paragraph{Sampling Process:}  
In each stratum \( s \), 
\begin{equation}
n_s = \Bigl\lfloor p \cdot |\text{Stratum}_s| \Bigr\rfloor
\end{equation}
samples are randomly selected. The full training set is the union of samples from all strata:
\begin{equation}
\mathcal{D}_{\text{train}} = \bigcup_{s=1}^{S} \mathrm{RandomChoose}\!\bigl(\text{Stratum}_s,\, n_s\bigr).
\end{equation}

\subsubsection{k-means Clustering Sampling}
\paragraph{Clustering Objective:}  
The k-means algorithm (with \( k = 7 \)) is applied to the measurement values \( v_i \) to minimize within-cluster variance:
\begin{equation}
J = \sum_{i=1}^{N} \|v_i - \mu_{c(i)}\|^2,
\end{equation}
where \( \mu_j \) is the centroid of cluster \( j \) and \( c(i) \) is the cluster label for sample \( i \).
\paragraph{Sampling Process:}  
After clustering into \( k \) clusters \( C_j \), each with \( |C_j| \) members, 
\begin{equation}
n_j = \Bigl\lfloor p \cdot |C_j| \Bigr\rfloor
\end{equation}
samples are randomly selected from each cluster. The training set is then formed as:
\begin{equation}
\mathcal{D}_{\text{train}} = \bigcup_{j=1}^{k} \mathrm{RandomChoose}\!\bigl(C_j,\, n_j\bigr).
\end{equation}

\section{Short Distance Elimination (SDE)}
\label{sec:SDE}
The SDE method eliminates points that are too close in the spatial coordinate system to previously selected points.

\paragraph{Distance Determination:}  
Define thresholds \( \alpha \) and \( \beta \) and the distance function as:
\begin{equation}
\Delta\!\bigl((x_i, y_i), (x_j, y_j)\bigr) =
\begin{cases}
1, & \text{if } |x_i - x_j| \ge \alpha \wedge |y_i - y_j| \ge \beta, \\
0, & \text{otherwise}.
\end{cases}
\end{equation}

\subsection{Sampling Process (Simplified):}

\begin{figure}[t]
    \centering
    \includegraphics[width=0.7\columnwidth]{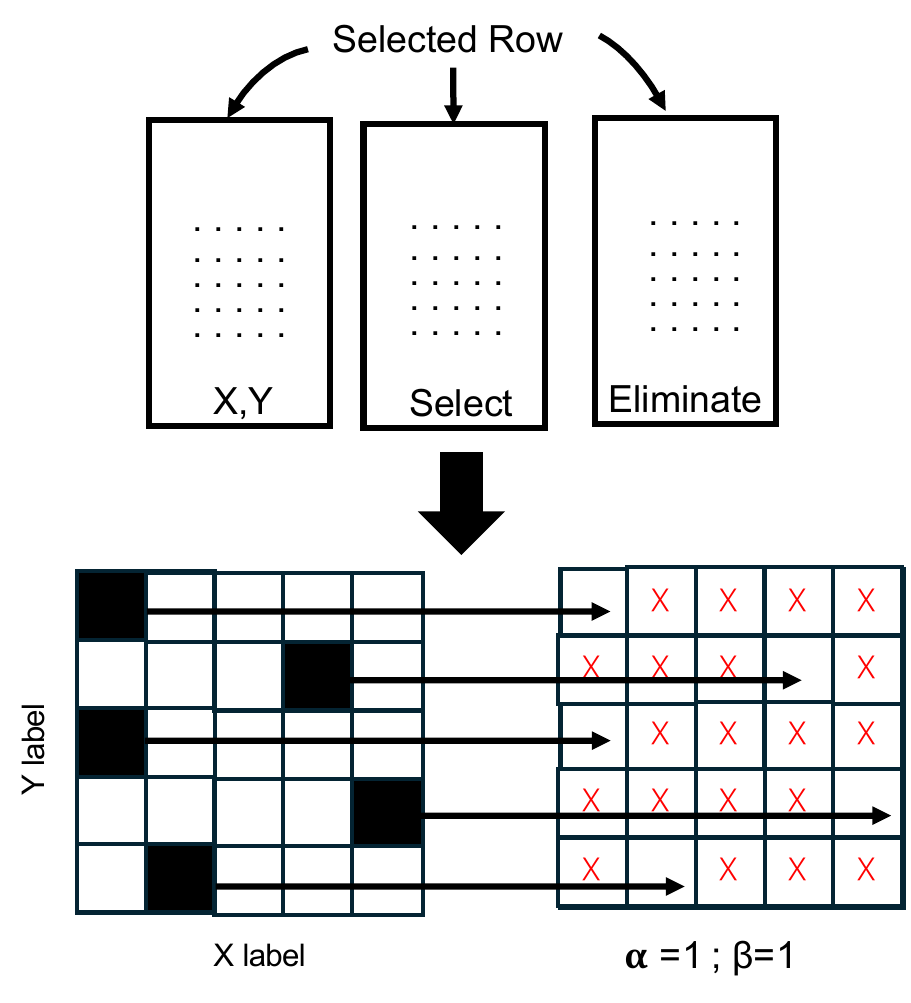}
    \caption{Sampling Process of Proposed SDE algorithm.}
    \label{fig:algorithm-ex}
\end{figure}

Figures~\ref{fig:algorithm-ex} illustrate the sampling workflow of the proposed Short Distance Elimination (SDE) algorithm.

In Figure~\ref{fig:algorithm-ex}, a 2D label matrix is shown where data points are either selected (black squares) or eliminated (red Xs) based on their spatial proximity. The algorithm uses threshold parameters \(\alpha\) and \(\beta\) set to one. However, customized values can also be specified to change the selection range of \(X\) and \(Y\).

\begin{enumerate}
    \item Initialize \( \mathrm{Selected} = \varnothing \).
    \item Randomly select an initial point and add it to \( \mathrm{Selected} \).
    \item Continue randomly selecting candidate points; if a candidate point satisfies \(\Delta = 1\) with respect to all already selected points, it is retained; otherwise, it is discarded.
    \item If the number of selected points is less than \(pN\), randomly select additional points from the discarded set until \(pN\) samples are obtained.
\end{enumerate}

\subsection{Parameter Sweep and Optimal (\(\alpha, \beta\))}
To determine the optimal \((\alpha, \beta)\) combination, \(\alpha\) and \(\beta\) were varied over the set \(\{0,1,2,3,4\}\) (with the restriction that they cannot both be zero), and the RMSD was computed for each combination. As shown in Table~\ref{tab:alpha_beta}, the best performance was achieved for \((\alpha,\beta) = (2,2)\) with the lowest RMSD of 0.0292. Accordingly, \((\alpha,\beta) = (2,2)\) is used in all subsequent experiments.

\subsection{Hybrid Methods: S-SDE and K-SDE}
By integrating SDE with stratified or clustering sampling, methods are obtained that leverage both the numerical distribution of measurement values and spatial dispersion.

\subsubsection{K-SDE: k-means + SDE}
\begin{figure}[t]
    \centering
    \includegraphics[width=0.9\columnwidth]{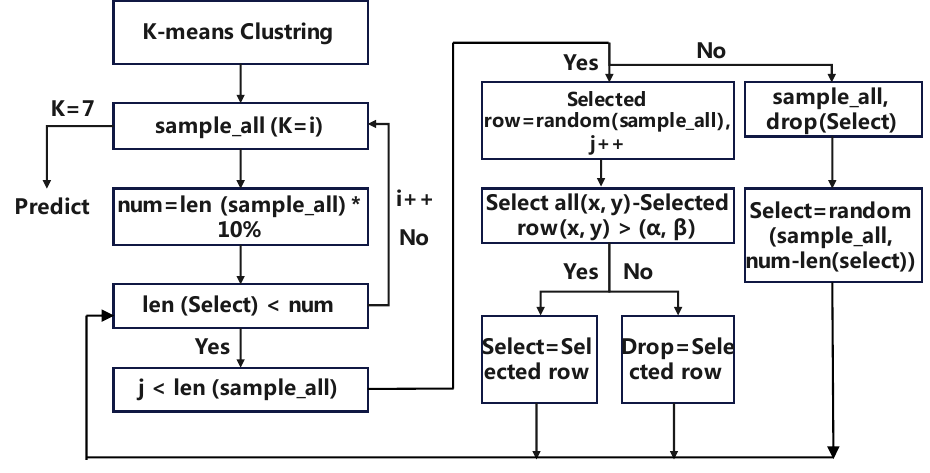}
    \captionsetup{font=footnotesize}
    \caption{Sampling Process of the Proposed SDE Algorithm with k-means Clustering.}
    \label{fig:sedexkmean}
\end{figure}

\paragraph{Clustering:}  
Partition the dataset into \(k\) clusters \(C_j\) using k-means based on measurement values. Figure~\ref{fig:sedexkmean} demonstrates how SDE is integrated with k-means clustering in the K-SDE strategy. First, data are grouped into clusters, then SDE is applied within each cluster to select spatially diverse points.

\paragraph{Applying SDE}  

Within each cluster \(C_j\), apply the SDE algorithm to select approximately
\begin{equation}
\lfloor p \cdot |C_j| \rfloor
\end{equation}
samples:
\begin{equation}
\mathcal{D}_j = \mathrm{SDE}\!\bigl(C_j,\; p\bigr).
\end{equation}

\paragraph{Global Training Set}  
The overall training set is the union of samples from all clusters:
\begin{equation}
\mathcal{D}_{\text{train}} = \bigcup_{j=1}^{k} \mathcal{D}_j.
\end{equation}

\subsubsection{S-SDE: Stratified + SDE}
\paragraph{Stratification:}  
Divide \(\mathcal{D}\) into \(S\) strata \(\mathrm{Stratum}_s\), each with \(|\mathrm{Stratum}_s|\) samples.

\paragraph{Applying SDE:}  
Within each stratum \(s\), apply the SDE algorithm to select approximately
\begin{equation}
\Bigl\lfloor p \times |\mathrm{Stratum}_s| \Bigr\rfloor
\end{equation}
spatially dispersed samples. Denote the selected samples in stratum \(s\) as:
\begin{equation}
\mathcal{D}_s = \mathrm{SDE}\!\bigl(\mathrm{Stratum}_s,\; p\bigr).
\end{equation}

\paragraph{Global Training Set:}  
The final training set is the union of samples from all strata:
\begin{equation}
\mathcal{D}_{\text{train}} = \bigcup_{s=1}^{S} \mathcal{D}_s.
\end{equation}

\section{Experimental Result} 
\label{sec:result}

\subsection{Datasets and Preprocessing}

\subsubsection{Wafer Data}
Experiments on wafer data were conducted using an industrial production test dataset of a \textbf{28 nm analog/RF} device. The dataset contains 23 wafers from one lot, each featuring approximately 6,000 devices under test (DUTs). A measured characteristic from the dynamic current test was utilized. A heat map of the full measurement results for the first wafer of the lot is displayed in Fig.~\ref{fig:wafer_heatmap}. For ease of experimentation, faulty dies were removed from the dataset. The dataset includes 16 sites in a single touchdown. However, site information was not considered separately during the training phase. Although the data originate from a multisite environment, all 16 sites were treated as a single entity because the source data did not exhibit discontinuous changes across sites.

\subsubsection{FPGA Data}
Experiments on FPGA data were conducted using measurements from the Xilinx Artix-7 FPGA. Data from 10 FPGAs (FPGA-01 to FPGA-05) were used. An on-chip measurement circuit was implemented using 7 stages (i.e., 7 LUTs) in a Configurable Logic Block (CLB). The Ring Oscillator (RO) is based on XNOR or XOR logic gates. By keeping the same internal routing, logic resources, and structure for each RO placed in a CLB, frequency variation caused by internal routing differences was minimized. A total of \textbf{3,173} ROs were placed on a geometrical grid of \(33 \times 120\) (excluding empty layout spaces) through hardware macro modeling using the Xilinx CAD tool, Vivado. Figure~\ref{fig:fpga_heatmap} shows a single path's measured frequency heat-map. This representation covers the total layout of the FPGA. The Artix-7 FPGA has 6-input LUTs, which create 32 (\(2^6 - 1\)) possible paths for XNOR- and XOR-based RO configurations (path-01 to path-32). In total, to complete the exhaustive fingerprint (X-FP) measurement, 32 fingerprint measurements were conducted for all 32 paths.

\subsection{Experimental Procedure}
\begin{enumerate}
    \item \textbf{Data Acquisition:} For each wafer or FPGA file, the data are read to obtain the dataset \( \mathcal{D} \).
    \item \textbf{Sampling:} Each of the five sampling methods (Random, Stratified, k-means, S-SDE, and K-SDE) is applied to select 10\% of \( \mathcal{D} \) as the training set.
    \item \textbf{Model Training:} A GPR model (using the RBF kernel) is trained on the selected training set; hyperparameters are determined via maximum likelihood estimation.
    \item \textbf{Prediction:} The trained model is used to predict the remaining 90\% of the data, and the RMSD is computed.
    \item \textbf{Result Recording:} The RMSD for each file is recorded and the average RMSD for wafer and FPGA datasets is computed.
\end{enumerate}

\subsection{Evaluation Metric}
The RMSD is computed as:
\begin{equation}
\mathrm{RMSD} = \sqrt{\frac{1}{|\mathcal{D}_{\text{test}}|} \sum_{(x_j,y_j,v_j)\in \mathcal{D}_{\text{test}}} \Bigl(\hat{v}_j - v_j\Bigr)^2}.
\end{equation}
A lower RMSD indicates higher prediction accuracy.

\subsection{Results and Discussion} \label{sec:results}

\subsection{Spatial Distribution Heatmaps}
\begin{figure}[t]
    \centering
    \includegraphics[width=0.7\columnwidth]{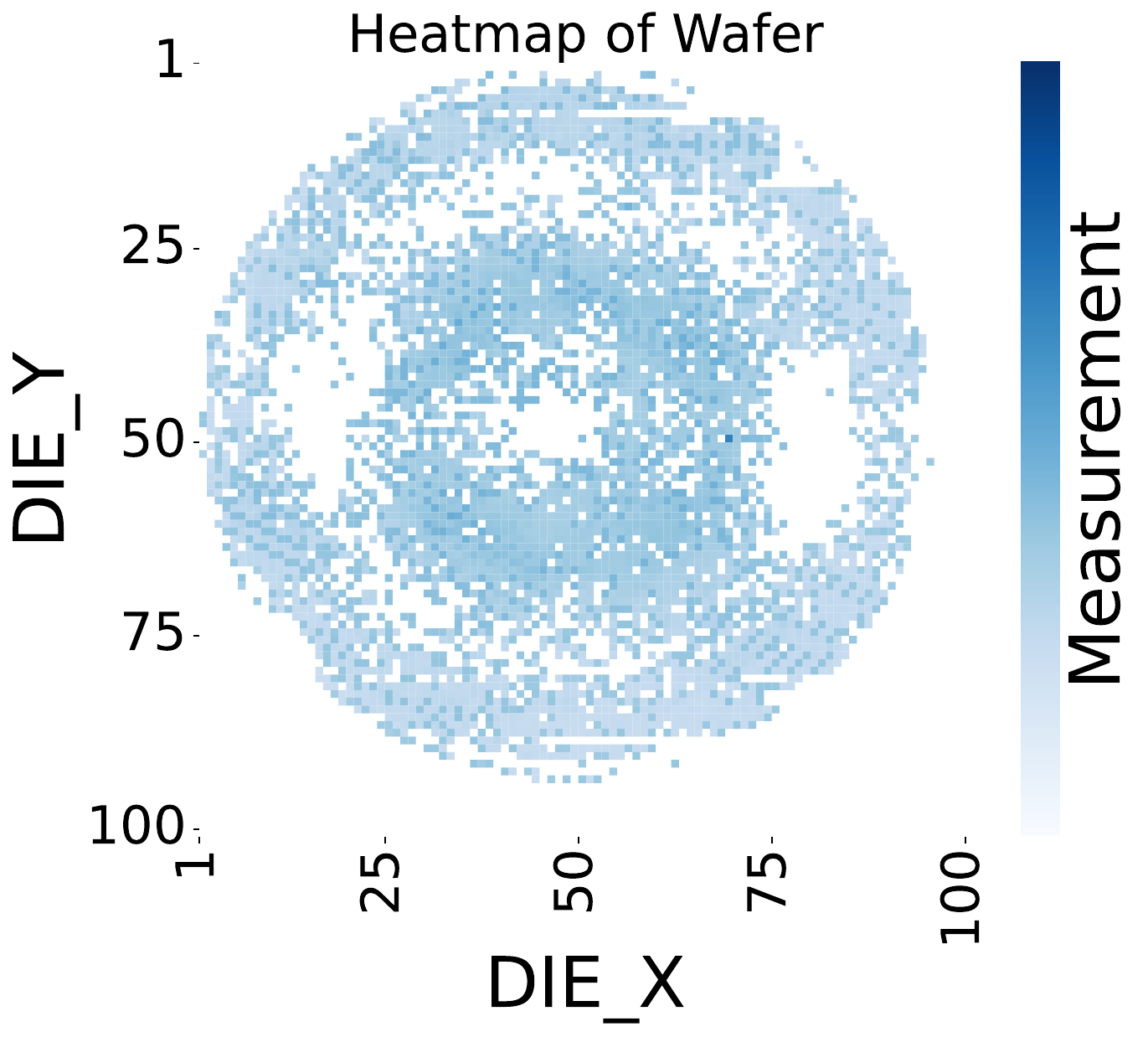}
    \captionsetup{font=footnotesize}
    \caption{Heatmap illustrating the spatial distribution in Wafer.}
    \label{fig:wafer_heatmap}
\end{figure}

\begin{figure}[t]
    \centering
    \includegraphics[width=0.65\columnwidth]{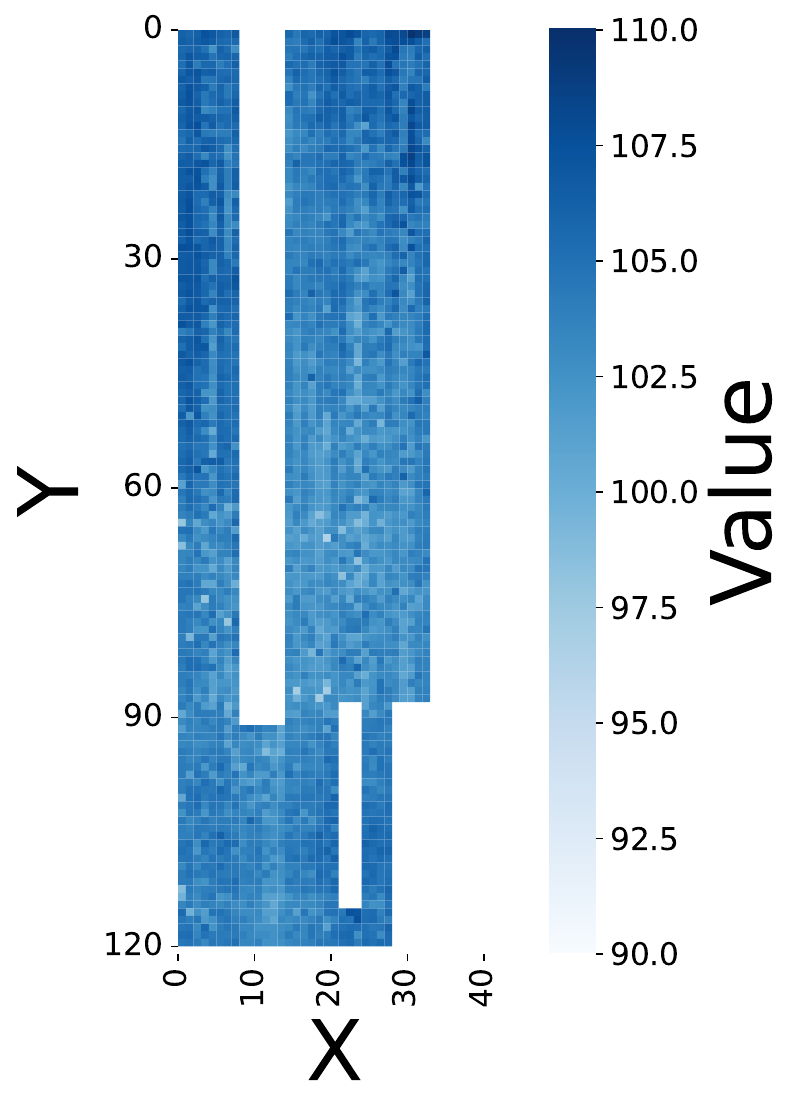}
    \captionsetup{font=footnotesize}
    \caption{Heatmap illustrating the spatial distribution in FPGA.}
    \label{fig:fpga_heatmap}
\end{figure}

Heatmaps are presented for both the Wafer and FPGA datasets. Figure~\ref{fig:wafer_heatmap} shows the spatial distribution in the Wafer data, while Figure~\ref{fig:fpga_heatmap} illustrates the spatial distribution in the FPGA data. These figures emphasize the importance of incorporating spatial information when selecting training samples.

\subsection{Effect of (\(\alpha,\beta\)) on SDE}
To determine the optimal \((\alpha,\beta)\) combination, \(\alpha\) and \(\beta\) were varied over the set \(\{0,1,2,3,4\}\) (with the restriction that they cannot both be zero), and the RMSD was computed for each combination. As shown in Table~\ref{tab:alpha_beta}, the best performance was achieved for \((\alpha,\beta) = (2,2)\) with the lowest RMSD of 0.0292. Accordingly, \((\alpha,\beta) = (2,2)\) is used in all subsequent experiments.

\begin{table}[ht]
  \scriptsize
  \centering
  \caption{Effect of \((\alpha,\beta)\) Parameter Variation on SDE RMSD (for one wafer data file)}
  \label{tab:alpha_beta} 
  \begin{tabular}{c|ccccc}
    \toprule
    \(\alpha\backslash\beta\) & 0      & 1      & 2        & 3      & 4      \\\midrule
    0 &   —    & 0.0301 & 0.0294   & 0.0298 & 0.0309 \\
    1 & 0.0299 & 0.0295 & 0.0301   & 0.0298 & 0.0304 \\
    2 & 0.0299 & 0.0299 & \textbf{0.0292} & 0.0294 & 0.0306 \\
    3 & 0.0304 & 0.0298 & 0.0305   & 0.0299 & 0.0300 \\
    4 & 0.0296 & 0.0299 & 0.0300   & 0.0295 & 0.0297 \\
    \bottomrule
  \end{tabular}
  \vspace{0.5ex}
  \begin{flushleft}
    \scriptsize\textit{Note: The lowest RMSD of 0.0292 is achieved for \((\alpha,\beta) = (2,2)\), which is used in all subsequent experiments.}
  \end{flushleft}
\end{table}

\subsection{Sampling Methods Comparison}
With the optimal \((\alpha,\beta)=(2,2)\) settings for SDE-based methods, the five sampling strategies were compared using both wafer and FPGA data. Table~\ref{tab:comparison} summarizes the average RMSD values (lower values indicate higher accuracy) for each sampling method.

\begin{table}[ht]
\caption{Performance Comparison of Sampling Methods}
\label{tab:comparison}
\centering
\begin{tabular}{@{}lcc@{}}
\toprule
Sampling Method & Wafer-RMSD  & FPGA-RMSD \\ \midrule
Random      & 0.307 & 1.976 \\
Stratified  & 0.050 & 0.985 \\
k-means     & 0.041 & 0.815 \\
S-SDE       & 0.042 & 0.898 \\
K-SDE       & 0.035 & 0.709 \\ \bottomrule
\end{tabular}
\end{table}

\textbf{Observations:}
\begin{itemize}
    \item Random Sampling yields the highest RMSD for both wafer and FPGA data.
    \item Stratified and k-means Sampling improve performance by ensuring adequate coverage of measurement value ranges.
    \item SDE-based methods further enhance performance by enforcing spatial dispersion; notably, the K-SDE method achieves the lowest RMSD in both cases.
\end{itemize}

Separate analyses were conducted for both datasets to assess consistency of the proposed methods.

\begin{figure}[t]
    \centering
    \includegraphics[width=0.99\columnwidth]{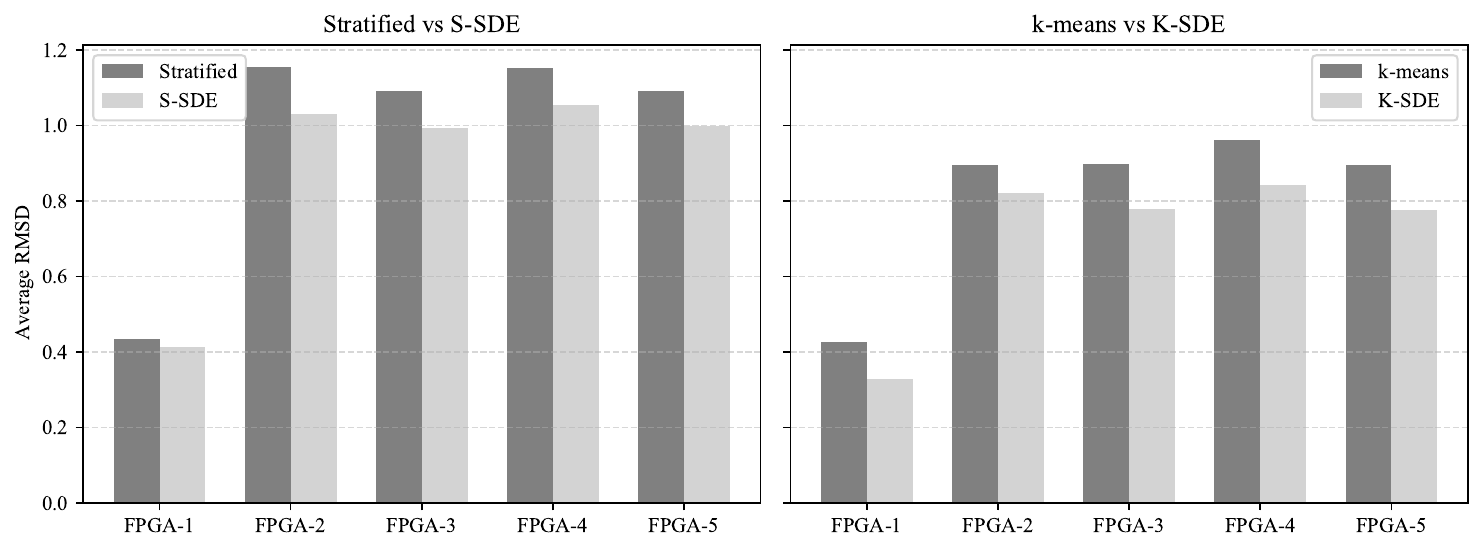}
    \captionsetup{font=footnotesize}
    \caption{Comparison with Baseline Sampling Methods across FPGAs.}
    \label{fig:fpgaavg}
\end{figure}

In the case of the FPGA lookup paths, as illustrated in Figure~\ref{fig:fpgaavg}, the proposed S-SDE and K-SDE consistently outperform their respective baselines (Stratified and k-means) across all five FPGAs.

\begin{figure}[t]
    \centering
    \includegraphics[width=0.99\columnwidth]{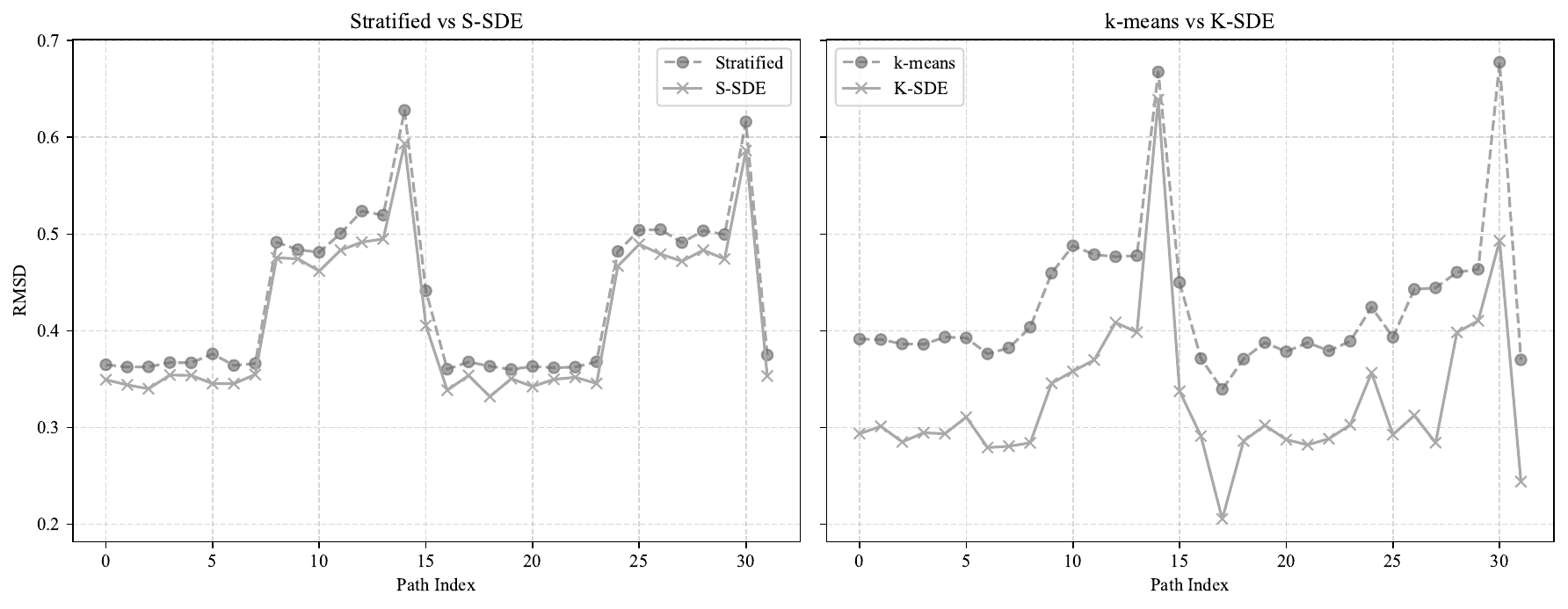}
    \captionsetup{font=footnotesize}
    \caption{Comparison with Baseline Sampling Methods Across All Paths of FPGA-1.}
    \label{fig:fpga-path}
\end{figure}

Additionally, path-wise performance within the first FPGA was examined, as shown in Figure~\ref{fig:fpga-path}. The results demonstrate that the proposed SDE algorithms outperform the baseline methods across nearly all 32 individual paths.

\begin{figure}[t]
    \centering
    \includegraphics[width=0.99\columnwidth]{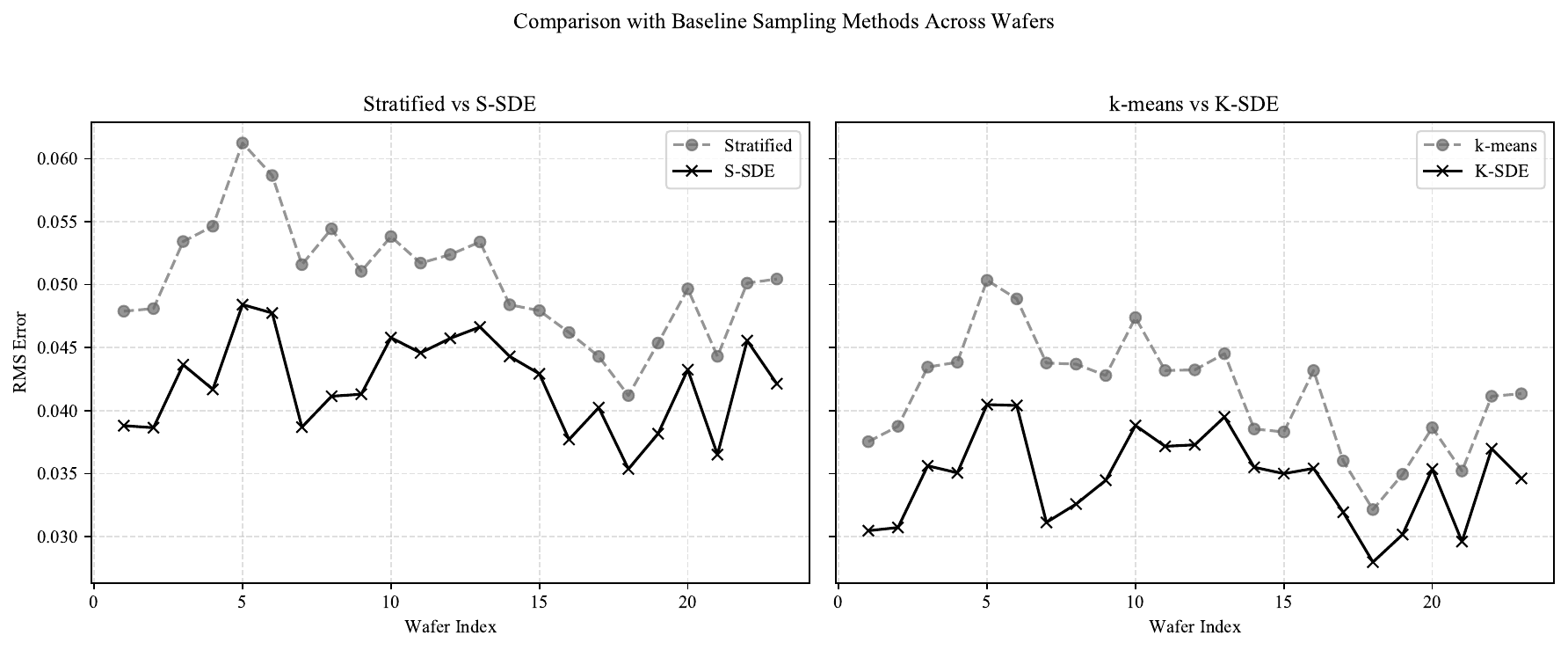}
    \captionsetup{font=footnotesize}
    \caption{Comparison with Baseline Sampling Methods Across Wafers.}
    \label{fig:waferdetail}
\end{figure}

Next, in Figure~\ref{fig:waferdetail}, improvement is evident in the average RMSD across all wafers within the lot. A minimum improvement of 8.47\% was observed in Wafer-14 and a maximum improvement of 25.01\% in Wafer-7 for S-SDE over Stratified. Similarly, K-SDE shows improvements over k-means ranging from 7.89\% (Wafer-14) to 28.91\% (Wafer-7).

Finally, the evaluation exhibits consistent improvement over the baseline, not only on average but also at the wafer level and path level. The results demonstrate that the proposed SDE method outperforms baseline techniques on both industry production wafer data and actual silicon-measured FPGA datasets.

\section{Conclusion and Future Work}\label{sec:conclusion}
High testing costs in semiconductor manufacturing have been addressed by selecting only 10\% of each test file’s data to train a Gaussian Process Regression (GPR) model and predict the remaining 90\%. Five sampling strategies were evaluated: Random, Stratified, k-means, and the proposed hybrid strategies S-SDE and K-SDE.

At the core of the approach is the Short Distance Elimination (SDE) algorithm, which discards spatially proximate candidate points to enforce spatial dispersion. The optimal configuration of SDE was identified as \((\alpha, \beta) = (2, 2)\) through a parameter sweep.

Experimental results demonstrate that:
\begin{itemize}
    \item Sampling methods based on measurement values (Stratified and k-means) outperform pure random sampling.
    \item Incorporating spatial dispersion information via SDE further enhances prediction performance.
    \item The K-SDE method—combining k-means clustering with SDE—achieves the best predictive accuracy for both wafer and FPGA datasets.
    \item K-SDE improves upon k-means sampling by 16.26\% for wafer data and 13.07\% for FPGA data.
    \item S-SDE improves upon stratified sampling by 16.49\% (wafer) and 8.84\% (FPGA).
\end{itemize}

Future research directions include applying the SDE approach to other machine learning models such as neural networks and random forests; developing efficient online sampling strategies for batch data environments; extending the method to support complex three-dimensional and large-scale spatial scenarios; and integrating manufacturing-specific constraints to enhance sampling refinement.

\bibliographystyle{IEEEtran}  % 让 BibTeX 用 IEEE 风格，且按“首次引用”排序
\bibliography{refs}           % “refs” 与你的 .bib 文件名一致

\end{document}